\documentclass{article}

\usepackage[preprint,nonatbib]{pr_class}

\usepackage{graphicx}
\usepackage[firstpage]{draftwatermark}
\SetWatermarkText{\hspace*{15cm}\raisebox{12.4cm}{\includegraphics{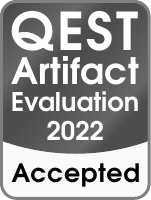}}}
\SetWatermarkAngle{0}
\usepackage{subfig}
\usepackage{microtype}
\usepackage{tipa}
\usepackage{mathtools}
\usepackage{fancyref}
\usepackage{float}
\restylefloat{figure}
\usepackage{amsmath}
\usepackage{amssymb}  
\usepackage{mathtools}
\usepackage{fancyhdr}
\usepackage{kantlipsum}
\usepackage{eso-pic}
\usepackage{nomencl}
\usepackage{etoolbox}
\usepackage{balance}
\usepackage{dsfont}
\usepackage{cite}
\usepackage{eucal}
\usepackage{nccmath}
\usepackage{tikz}
\usepackage{xcolor}
\tikzset{>=latex}
\usetikzlibrary{fit,automata,positioning,angles,quotes}
\usepackage[algo2e,linesnumbered,ruled,lined]{algorithm2e}
\usepackage[toc,page]{appendix}
\usepackage{hyperref}
\hypersetup{
	colorlinks=true,
	linkcolor=black,
	filecolor=black,      
	urlcolor=black,
	citecolor=black,}
\usepackage[T1]{fontenc}
\usepackage{lmodern}
\usepackage{multirow}
\usepackage{diagbox}

\usepackage[utf8]{inputenc}

\usepackage{listings}
\definecolor{textblue}{rgb}{.2,.2,.7}
\definecolor{textred}{rgb}{0.54,0,0}
\definecolor{textgreen}{rgb}{0,0.43,0}
\newtheorem{remark}{\textbf{Remark}}

\newcommand{\crv}{\textcolor{black}}
\newcommand{\lcrl}{{\ttfamily LCRL}\xspace}
\title{{\sffamily \bfseries LCRL}: Certified Policy Synthesis via\\ Logically-Constrained Reinforcement Learning\thanks{This work is in part supported by the HiClass project (113213), a partnership between the Aerospace Technology Institute (ATI), Department for Business, Energy and Industrial Strategy (BEIS) and Innovate UK.}
}

\author{Hosein~Hasanbeig,
	Daniel~Kroening\thanks{The work in this paper was done prior to joining Amazon.} ,
	Alessandro~Abate\\
Computer Science Department, University of Oxford, UK,\\ Amazon Inc, London, UK\\
	{\{hosein.hasanbeig, alessandro.abate\}@cs.ox.ac.uk, daniel.kroening@magd.ox.ac.uk}}

\begin{document}
\maketitle

\begin{abstract}
\lcrl is a software tool that implements model-free Reinforcement Learning (RL) algorithms 
over unknown Markov Decision Processes (MDPs), 
synthesising policies that satisfy a given linear temporal specification with maximal probability.  
\lcrl leverages partially deterministic finite-state machines known as Limit Deterministic B\"uchi Automata (LDBA)  
to express a given linear temporal specification. 
A reward function for the RL algorithm is shaped on-the-fly, based on the structure of the LDBA.  
Theoretical guarantees under proper assumptions ensure the 
convergence of the RL algorithm to an optimal policy that maximises the satisfaction probability.  
We present case studies to 
demonstrate the applicability, ease of use, scalability and performance of \lcrl~. Owing to the LDBA-guided exploration
and \lcrl model-free architecture, we observe robust performance, which also scales well when compared to standard RL approaches (whenever applicable to LTL specifications). Full instructions on how to execute all the case studies in this
paper are provided on a GitHub page that accompanies the \lcrl
distribution \url{www.github.com/grockious/lcrl}.

\end{abstract}

\section{Introduction}
%Over the past few years, Reinforcement Learning (RL) has attracted unprecedented recognition and achieved impressive results in a wide range of decision-making problems. This success is mainly owed to its ability in inferring control policies directly from data, without any prior knowledge of the problem. 
%\edit{However, the learned control policy is often difficult to understand and examine by humans, and the robustness and reproducibility of RL results are usually evaluated by statistical testing [(3) it is not clear how our approach solves these issues.]}. 
%\edit{The principal reason for this lack of insight is perhaps [(4) this is arguable]} RL objective to solely maximise the expected reward, i.e. a plain scalar feedback function. 
%This essentially means that there is no systematic method to guarantee that a solution synthesised using RL meets the expectations of the designer of the reward function or the learning algorithm. When applying RL to safety-critical systems, this becomes a pressing issue and there is an evident need for RL algorithms that are designed to learn human-interpretable controllers with formal guarantees.

Markov Decision Processes (MDPs) are extensively used for problems in which an agent needs to control a process by selecting actions that are allowed at the process' states and that affect state transitions. Decision making problems in MDPs are equivalent to resolving action non-determinism, and result in policy synthesis problems. Policies are synthesised to maximise expected long-term rewards obtained from the process. 
This paper introduces a new software tool, \lcrl, which performs policy synthesis for unknown MDPs when the goal is that of maximising the probability to abide by a task (or constraint) that is specified using Linear Temporal Logic (LTL).  
LTL is a formal, high-level, and intuitive language to describe complex tasks~\cite{clarke}.  
In particular, unlike static (space-dependent) rewards, LTL can describe time-dependent and complex non-Markovian tasks that can be derived from natural languages~\cite{natural2LTL,natural2LTL2,natural2LTL3}. 
%LTL has been employed in a number of industrial applications such as robotics, safe planning, and complex networks management. 
Any LTL specification can be translated efficiently into a Limit-Deterministic B\"uchi Automaton (LDBA), which allows \lcrl to automatically shape a reward function for the task that is later employed by the RL learner for optimal policy synthesis.  
\lcrl is implemented in \texttt{Python}, the \emph{de facto} standard programming language
for machine learning applications. 
%This makes \lcrl a good starting point for future work by other researchers.

\subsection{Related Work}

There exists a few tools that solve control (policy) synthesis in a model-free fashion, but not under full LTL specifications.  
One exception is the work in \cite{bozkurt} which proposes an interleaved reward and discounting mechanism. However, the reward shaping dependence on the discounting mechanism can make the reward sparse and small, 
which might negatively affect convergence. The work in \cite{hahn2021mungojerrie} puts forward a tool for an average-reward scheme based on earlier theoretical work.  
Other model-free approaches with available code-bases are either (1) focused on fragments of LTL and classes of regular languages (namely finite-horizon specs) or (2) cannot deal with unknown black-box MDPs.  
The proposed approach in \cite{qrm,jothim} presents a model-free RL solution but for regular specifications that are expressed as deterministic finite-state machines. The work in
\cite{bolt,bolts_repo} takes a set of LTL${}_f$/LDL${}_f$ formulae interpreted over finite traces as constraints, and then finds a policy that maximises an external reward function.  
The \texttt{VSRL} software tool \cite{vsrl,fulton,fulton2,fulton3} solves a control synthesis problem whilst maintaining a set of safety constraints during learning. COOL-MC~\cite{gross2022cool}, an open-source tool, integrates the OpenAI gym
with the probabilistic model checker Storm~\cite{storm}.

\subsection{Contributions}

%\hos{A reviewer asked to elaborate on what the inputs are, what the outputs, and the mapping between them that the tool performs. I think this is a reasonable request. Use Fig 1 better?} \green{This has been elaborated on in Section 2.2, and 2.3.}

\begin{figure}[!h]
	\centering
	\includegraphics[width=0.9\textwidth]{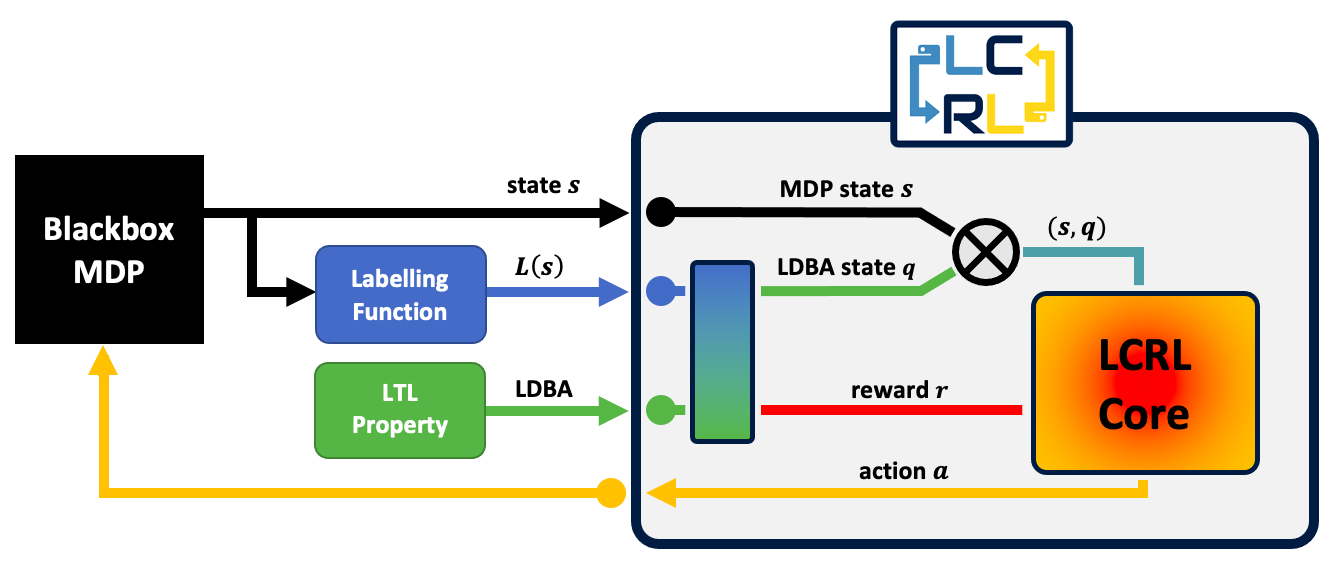}
	\caption{The \lcrl architecture: the inputs to the tool are the environment (MDP) - in particular its states $s$ and corresponding labels $L(s)$ - as well as the LDBA generated from the user-defined LTL specification. The MDP (with state $s$) and the LDBA (with state $q$) are synchronised on-the-fly, thus generating the pair $(s,q)$. A reward $r$ is then automatically generated by \lcrl, based on the environment label $L(s)$ and the automaton state $q$; actions $a$ are selected accordingly and passed back to the environment (MDP). } 
	\label{fig:overview}
\end{figure}

The \lcrl software tool has the architecture presented in Figure \ref{fig:overview}, and presents the following features: 
\begin{itemize}

\item \lcrl leverages \textbf{model-free RL algorithms}, employing only traces of the system
(assumed to be an unknown MDP) to formally synthesise optimal policies that
satisfy a given LTL specification with maximal probability.  \lcrl finds such
policies by learning over a set of traces extracted from the MDP under
LTL-guided exploration. \crv{This efficient, guided exploration is owed to reward shaping based on the automaton  \cite{lcrl_arxiv,lcnfq,plmdp,lcrl_j,hasanbeig2020safe,deepsynth}.} The guided exploration enables the algorithm to
focus only on relevant parts of the state/action spaces, as opposed to
traditional Dynamic Programming (DP) solutions, where the Bellman iteration is
exhaustively applied over the whole state/action spaces \cite{NDP}.  Under standard 
RL convergence assumptions, the \lcrl output is an optimal
policy whose traces satisfy the given LTL specification with \textbf{maximal} 
probability.

\item \lcrl is \textbf{scalable} owing to LTL-guided exploration, which allows
\lcrl to cope and perform efficiently with MDPs whose state and action
spaces are significantly large. There exist a few LDBA construction algorithms for LTL, but
not all of resulting automata can be employed for quantitative
model-checking and probabilistic synthesis \cite{kini}. The succinctness of the construction proposed in \cite{sickert}, 
which is used in \lcrl, is another contributing factor to \lcrl scalability. The
scalability of \lcrl is evaluated in an array of numerical examples
and benchmarks including high-dimensional Atari 2600 games
\cite{arcade,gym}.

\item \lcrl is the first RL synthesis method for LTL specifications in \textbf{continuous
state/action} MDPs. So far no tool is available to enable RL, whether
model-based or model-free, to synthesise policies for LTL on continuous-state/action MDPs.  
%\lcrl is \textbf{the first} method in this area, against which we compared the most well-known methods to deal with continuous spaces \cite{voronoi_o,voronoi,gordon,reachability_in_hybrid,faust,stochy}.  More specifically, 
Alternative approaches for continuous-space MDPs \cite{voronoi_o,voronoi,reachability_in_hybrid,faust} discretise the model into a finite-state MDP, or alternatively propose a DP-based method with value function approximation \cite{gordon}. 

\item \lcrl displays \textbf{robustness} features to hyper-parameter tuning. 
Specifically, we observed that \lcrl results, although
problem-specific, are not significantly affected when hyper-parameters are
not tuned with care.  %This is arguably due to the fact that the number of hyper-parameters to tune in RL is smaller when an expected discounted return is used, as compared to other objective formalisations in RL.

\end{itemize} 

%The rest of the paper is structured as follows: Section~\ref{sec:theory} presents a very brief review of the theoretical underpinnings of the \lcrl tool. Implementation and functionality of \lcrl is detailed in Section~\ref{sec:overview}. Finally, we highlight features and showcase the performance of \lcrl by a set of experimental evaluations in Section~\ref{sec:experiments}. 

%\section{Brief Overview of The Theory}\label{sec:theory}   

\section{Logically-Constrained Reinforcement Learning (LCRL)}\label{sec:overview}

We assume the unknown environment is encompassed by an MDP, 
which in this work is a discrete-time stochastic control processes defined as the tuple  $\mathfrak{M}=(\mathcal{S},\allowbreak\mathcal{A},s_0,\allowbreak
P)$ over a set of continuous states $\mathcal{S}=\mathbb{R}^n$, 
and where $\mathcal{A}=\mathbb{R}^m$ is a set of continuous actions, 
and $s_0 \in \mathcal{S}$ is an initial state.
$P:\mathcal{B}(\mathcal{S})\times\mathcal{S}\times\mathcal{A}\rightarrow [0,1]$ is a conditional transition kernel which assigns to any pair comprising a state $s \in \mathcal{S}$ and an action $a \in \mathcal{A}$ a probability measure $P(\cdot|s,a)$ on 
%$(\mathcal{S},\mathcal{B}(\mathcal{S}))$, where $\mathcal{B}(\mathcal{S})$ is the set of all Borel 
properly measurable sets on $\mathcal{S}$ \cite{shreve}. 
A finite state/action MDP is a special case in which $|\mathcal{S}|<\infty,~|\mathcal{A}|<\infty$, and $P:\mathcal{S}\times\mathcal{A}\times\mathcal{S}\rightarrow[0,1]$ is a transition probability matrix assigning a conditional probability to enter sets of states in $\mathcal{S}$.  
A variable $R(s,a)\sim\varUpsilon(\cdot|s,a)\in\mathcal{P}(\mathbb{R^+})$ is defined over the MDP~$\mathfrak{M}$, representing the reward obtained when action $a$ is taken at a given state~$s$, 
sampled from the reward distribution $\varUpsilon$ defined over the set of probability distributions $\mathcal{P}(\mathbb{R^+})$ on subsets of $\mathbb{R^+}$.  

\medskip 

\lcrl is a policy synthesis architecture for tasks that are expressed as specifications in LTL \cite{lcrl_arxiv,lcnfq,plmdp,lcrl_j,overlay,cautiousRL,hasanbeig2020deep,hasanbeig2020safe,deepsynth}.  
The \texttt{LCRL Core} in Fig.~\ref{fig:overview} is compatible
with any general RL scheme that conforms with the environment state and action spaces.  
Inside the \lcrl module the MDP and LDBA states are synchronised, resulting in an on-the-fly product MDP.
Intuitively, the product MDP encompasses the extra dimension of the LDBA states, 
which is added to the state space of the original MDP.  The role of
the added dimension is to track the sequence of labels that have been read across episodes,  
and thus to act as a memory register for the given task. This allows to evaluate the (partial) satisfaction
of the corresponding temporal property. 
More importantly, this synchronisation breaks down the non-Markovian LTL specification into a set of Markovian reach-avoid components, 
%over the product MDP 
which facilitates the RL convergence to a policy whose traces satisfy the overall LTL specification. 
In practice, no product between the MDP and LDBA is computed: the LDBA simply \emph{monitors} traces executed by the agents as episodes of the RL scheme.

\begin{remark}\label{remark:epsilon} 
%\hos{[This remark is still quite unclear, as it delves into the details of the non-determinism of the LDBA which has not been given formalisation. Unless this epsilon non-determinism appears explicitly elsewhere in this manuscript, I would recommend reducing this discussion to bare mention and add cit to tech references for more details.] }
%
The LDBA construction inherently introduces limited form of non-determinism, called $\varepsilon$-transitions,
which is treated as an extra action over the original MDP action space \cite{sickert,sickert2}.  
Namely, when there exists a non-deterministic transition in an LDBA state, 
the MDP action space is augmented with the non-deterministic
transition predicate of the LDBA. 
These non-deterministic transitions
are automatically handled by \lcrl during the learning and appropriate
on-the-fly modifications are carried out,  
%\hos{[is it necessary to know these details, for the sake of the tool?]} This also affects the product-MDP action space in the same manner and allows 
so that the RL agent can learn to deal with those non-deterministic transitions in order to reach
the accepting conditions of the LDBA. 
% \edit{[so far not very clear - what is important are the next two sentences:]} 
We emphasise that the underlying assumption in \lcrl is that the MDP model is unknown (Fig.~\ref{fig:overview}), and thus a single state is obtained as output when given a state and an action as input.
\hfill$\square$
\end{remark}

\lcrl defines a reward function $R$ for the \texttt{RL Core}, whose objective is to maximise the expected discounted
return~\cite{sutton}: 
%\hos{[the use of maths symbols is abrupt: we do not discuss the space of the state/action variables - a few details of the MDP are in the next session. The reward should be introduced as a function proper - we need to find a better balance. ]}\green{fixed} 
\begin{align}\label{upol}
\begin{aligned}
&\mathds{E}^\pi[\sum\limits_{n=0}^{\infty} \gamma^n~ R(s_n,a_n)|s_0=s],
\end{aligned}
\end{align}
where $\mathds{E}^{\pi} [\cdot]$ denotes the expected value given that the agent follows the policy $\pi:\mathcal{S} \times \mathcal{A} \rightarrow [0,1]$ from state $ s $; 
parameter $\gamma\in [0,1]$ is a discount factor; 
and $s_0,a_0,s_1,a_1...$ is the sequence of state/action pairs, initialised at $s_0 = s$.  
This reward is intrinsically defined over the product MDP, namely it is a function of the MDP state (describing where the agent is in the environment) and the sate of the automaton (encompassing partial task satisfaction). \crv{For further details on the \lcrl reward shaping, please refer to \cite{lcrl_arxiv,lcnfq,plmdp,lcrl_j,hasanbeig2020safe,deepsynth}}.  

%\hos{[this section requires much more work: we should try to exclusively introduce the maths that is needed for the user of the tool - essentially we should project all the technical and algorithmic details onto this plane - can we align this with the details (e.g., hyper-params choices) in the next section? The remaining technical aspects should be discussed uniquely at a qualitative level. ]} \green{fixed} 

The discount factor $\gamma$ affects the optimality of the synthesised
policy and has to be tuned with care.  
There is standard work in RL on state-dependent discount factors
\cite{discount,discount2,discount3,discount4,bozkurt}, which is shown
to preserve convergence and optimality guarantees.  
For \lcrl the learner discounts the received reward whenever it's positive, and leaves it un-discounted otherwise:
\begin{equation}\label{eq:gamma_in}
\gamma(s) = \left\{
\begin{array}{ll}
\eta & $ if $ R(s,a)>0,\\
1 & $ otherwise, 
$
\end{array}
\right.
\end{equation}
where $0<\eta<1$ is a constant \cite{discount2,lcrl_j}. Hence, \eqref{upol} reduces to an expected return that is bounded, namely  
\begin{equation}\label{state_dep_utility_in}
\mathds{E}^{\pi} [\sum\limits_{n=0}^{\infty} \gamma(s_n)^{N(s_n)}~ R(s_n,\pi(s_n))|s_0=s],~0<\gamma(s)\leq 1,
\end{equation}
where $N(s_n)$ is the number of times a positive reward has been observed at state~$s_n$. 

For any state $s \in \mathcal{S}$ and any action $a \in \mathcal{A}$, \lcrl assigns a quantitative action-value $Q:\mathcal{S}\times\mathcal{A}\rightarrow \mathbb{R}$, which is initialised with an arbitrary and finite value over all state-action pairs. As the agent starts learning, the action-value $Q(s,a)$ is updated by a linear combination between the current $Q(s,a)$ and the target value: $$R(s,a)+\gamma \max\limits_{a' \in \mathcal{A}}Q({s'},a'),$$with the weight factors $1-\mu$ and $\mu$ respectively, where $\mu$ is the learning rate. 

An optimal stationary Markov policy synthesised by \lcrl on the product MDP
that maximises the expected return, is guaranteed to induce a finite-memory
policy on the original MDP that maximises the probability of satisfying the
given LTL specification \cite{lcrl_j}.  Of course, in finite-state and -action MDPs, the set
of stationary deterministic policies is finite and thus after a finite
number of learning steps RL converges to an optimal policy.  However, when
function approximators are used in RL to tackle extensive or even infinite-state (or
-action) MDPs, such theoretical guarantees are valid only asymptotically \cite{lcnfq,hasanbeig2020deep}.

\subsection{Installation}
\lcrl can be set up by the \texttt{pip} package manager as easy as: 
\begin{center}
	\texttt{pip install lcrl}
\end{center}
This allows to readily import \lcrl as a package into any \texttt{Python} project  
\begin{center}
	\texttt{$>>>$ import lcrl}
\end{center}
and employ its modules.  Alternatively, the provided \texttt{setup} file
found within the distribution package will automatically install all the
required dependencies.  The installation setup has been tested successfully
on Ubuntu~18.04.1, macOS~11.6.5, and Windows~11.

\subsection{Input Interface}
\lcrl training module \texttt{lcrl.src.train} inputs two main objects (cf. Fig. \ref{fig:overview}):  
an \textbf{MDP} black-box object that generates training episodes; 
and an \textbf{LDBA} object; 
as well as learning hyper-parameters\footnote{These parameters are called hyper-parameters since their values are used to control the learning process. This is unlike other parameters, such as weights and biases in neural networks, which are set and updated automatically during the learning phase.} that are listed in
Table~\ref{tab:hyper_parameters}.

\bgroup
\def\arraystretch{1.25}
\begin{table}[!t]
\centering
\caption{List of hyper-parameters and features that can be externally selected}
\label{tab:hyper_parameters}
\vspace{2mm}
\resizebox{\textwidth}{!}{%
	\begin{tabular}{|l|c|l|}
		\hline
		\textbf{\colorbox{white}{Hyper-parameter}} & \textbf{~Default Value~} & \textbf{~Description} \\ \hline
		\multirow{4}{*}{~\texttt{algorithm}} & \multirow{4}{*}{`ql'} & \multirow{4}{*}{\begin{tabular}[l]{@{}l@{}l@{}l@{}}~RL algorithm underlying \texttt{LCRL Core}, selected between (cf. Tab. \ref{tab:results}):\\
				-~`ql': Q-learning, \\
				-~`nfq': Neural Fitted Q-iteration, \\
				-~`ddpg': Deep Deterministic Policy Gradient\end{tabular}}\\
		& & \\
		& & \\
		& & \\
		~\texttt{episode\_num} & 2500 & ~number of learning episodes \\ 
		~\texttt{iteration\_num\_max}~ & 4000 & ~max number of iterations/steps within each episode \\ 
		~\texttt{discount\_factor} & 0.95 & ~discounting coefficient $\eta$ as in \eqref{eq:gamma_in} \\ 
		~\texttt{learning\_rate} & 0.9 & ~learning rate parameter $\mu$ \\ 
		~\texttt{epsilon} & 0.1 & ~value for epsilon-greedy exploration ($= 0$ for fully greedy) \\ 
		~\texttt{test} & true & ~run of closed-loop simulations to test the generated policy \\ 
		~\texttt{save\_dir} & `./results' & ~directory address for saving the results \\ 				
		\multirow{2}{*}{~\texttt{average\_window}} & \multirow{2}{*}{-1} & \multirow{2}{*}{\begin{tabular}[l]{@{}l@{}}~number of episodes for moving-average window for plots \\ (default value -1 for 30\% of episode\_num)\end{tabular}} \\
		&  &  \\ \hline
	\end{tabular}%
}
\vspace{-4mm}
\end{table}
\egroup

\subsubsection{MDP:}
An MDP is an object with internal attributes that are a priori unknown to the agent, namely
the state space, 
the transition kernel, 
and the labelling function (respectively denoted by $\mathcal{S}$, $P$, and $L$). The states and their labels are observable upon reaching.
To formalise the agent-MDP interface we adopt a scheme that is widely accepted in the RL literature
\cite{gym}. In this scheme the learning agent can invoke the following methods from any state of the MDP:
\begin{itemize}

\item \textbf{\texttt{reset()}}: this resets the MDP to its initial state. This allows the agent to start a new learning episode whenever necessary.

\item \textbf{\texttt{step(action)}}: the MDP \texttt{step} function takes an
action (the yellow signal in Fig.~\ref{fig:overview}) as input, and
outputs a new state, i.e. the black signal in Fig.~\ref{fig:overview}.

\end{itemize}
A number of well-known MDP environments (e.g., the stochastic grid-world) are embedded as classes within \lcrl, and can be found within
the module \texttt{lcrl.src.environments}.  
Most of these classes can easily set up an MDP object. However, note that the
state signal output by the \texttt{step} function needs to be fed to a
labelling function \texttt{state\_label(state)}, which outputs a
list of labels (in string format) for its input \texttt{state}
(in Fig.~\ref{fig:overview}, the black output signal from the MDP is fed to the blue box, or labelling function, which outputs the set of label).  For example,
\texttt{state\_label(state)=[`safe', `room1']}. 
%The important point is
%that the labelling function is not available to the learning agent, which instead  
%to be called for any state and at any time step.  The agent only 
%receives the label of the state from the MDP.  
The labelling function \texttt{state\_label(state)} can then be positioned
outside of the MDP class, or it can be an internal method in the MDP class. 
The built-in MDP classes in \texttt{lcrl.src.environments} module have an empty
\texttt{state\_label(state)} method that are ready to be overridden at the instance level:

\
\begin{lstlisting}[caption={Example of \texttt{state\_label(state)} specification in the MDP object \texttt{lcrl.src.environments.gridworld\_1}.},captionpos=b]
# create a SlipperyGrid object
gridworld_1 = SlipperyGrid()

# "state_label" function outputs the label of input state 
# (input: state, output: string label)
def state_label(self, state):
	# defines the labelling image
	labels = np.empty([gridworld_1.shape[0], gridworld_1.shape[1]], dtype=object)
	labels[0:40, 0:40] = 'safe'
	labels[25:33, 7:15] = 'unsafe'
	labels[7:15, 25:33] = 'unsafe'
	labels[15:25, 15:25] = 'goal1'
	labels[33:40, 0:7] = 'goal2'
	# returns the label associated with input state
	return labels[state[0], state[1]]
		
# now override the step function
SlipperyGrid.state_label = state_label.__get__(gridworld_1, SlipperyGrid)
\end{lstlisting}

\subsubsection{LDBA:}
An LDBA object is an instance of the \texttt{lcrl.src.automata.ldba} class. 
This class is structured according to the automaton construction in \cite{sickert}, and it encompasses modifications dealing with non-determinism, as per Remark~\ref{remark:epsilon}. 
%An LDBA is a finite-state machine where the states are numbered, i.e.  there is a mapping from $\mathcal{Q}$ to $\mathbb{N}_0$.  
The LDBA initial state is numbered as $0$, or can alternatively be specified using the class attribute \texttt{initial\_automaton\_state} once an LDBA object is created. The LDBA non-accepting sink state is numbered as $-1$. Finally, the set of accepting sets, on which we elaborate further below, has to be specified at the instance level by configuring \texttt{accepting\_sets} (Listing~\ref{fig:ldba_step} line 1).
The key interface methods for the LDBA object are:
\begin{itemize}
	\item \textbf{\texttt{accepting\_frontier\_function(state)}}: this automatically updates an internal attribute of an LDBA class called \texttt{accepting\_sets}. This is a list of accepting sets of the LDBA, e.g. $\mathcal{F}=\{F_1,...,F_f\}$. For instance, if the set of LDBA accepting sets is $\mathcal{F}=\{\{3,4\},~\{5,6\}\}$ then this attribute is a list of corresponding state numbers \texttt{accepting\_sets=[[3,4],[5,6]]}. As discussed above, the \texttt{accepting\_sets} has to be specified once the LDBA class is instanced (Listing~\ref{fig:ldba_step} line 1). 
	The main role of the accepting frontier function is to determine if an accepting set can be reached, so that a corresponding reward is given to the agent (cf. red signal in Fig.~\ref{fig:overview}). 
	%[remaining part of paragraph: clearly not self contained, unless you've read all the theory elsewhere:] 
	Once an accepting set is visited it will be temporarily removed from the \texttt{accepting\_sets} until the agent visits all the accepting sets within \texttt{accepting\_sets}. After that, \texttt{accepting\_sets} is reset to the original list. \crv{To set up an LDBA class in \lcrl the user needs to specify \texttt{accepting\_sets} for the LDBA. \lcrl then automatically shapes the reward function and calls the \texttt{accepting\_frontier\_function} whenever necessary. Further details on the \texttt{accepting\_frontier\_function} and the \texttt{accepting\_sets} can be found in \cite{lcrl_arxiv,lcnfq,plmdp,lcrl_j,hasanbeig2020safe,deepsynth}.}
	
	\item \textbf{\texttt{step(label)}}: LDBA \texttt{step} function takes a label set, i.e. the blue signal in Fig.~\ref{fig:overview}, as input and outputs a new LDBA state. The label set is delivered to the \texttt{step} function by \lcrl. The \texttt{step} method is empty by default and has to be specified manually after the LDBA class is instanced (Listing.~\ref{fig:ldba_step} line 5).    
	\item \textbf{\texttt{reset()}}: this method resets the \texttt{state} and \texttt{accepting\_sets} to their initial assignments. This corresponds to the agent starting a new learning episode. 
\begin{lstlisting}[caption={Example of the specification of the \texttt{step(label)} method in the LDBA object \texttt{lcrl.automata.goal1\_or\_goal2} for the LTL specification $(\lozenge \square \mathit{goal1} \vee \lozenge \square \mathit{goal2}) \wedge \square \neg \mathit{unsafe}$. The non-accepting sink state is numbered as $-1$.},captionpos=b,label={fig:ldba_step}]
	goal1_or_goal2 = LDBA(accepting_sets=[[1, 2]])
	
	# "step" function for the automaton transitions 
	# (input: label, output: automaton_state, non-accepting sink state is "-1")
	def step(self, label):
	# state 0
	if self.automaton_state == 0:
	if 'epsilon_1' in label:
	self.automaton_state = 1
	elif 'epsilon_2' in label:
	self.automaton_state = 2
	elif 'unsafe' in label:
	self.automaton_state = -1  # non-accepting sink state
	else:
	self.automaton_state = 0
	# state 1
	elif self.automaton_state == 1:
	if 'goal1' in label and 'unsafe' not in label:
	self.automaton_state = 1
	else:
	self.automaton_state = -1  # non-accepting sink state
	# state 2
	elif self.automaton_state == 2:
	if 'goal2' in label and 'unsafe' not in label:
	self.automaton_state = 2
	else:
	self.automaton_state = -1  # non-accepting sink state
	# step function returns the new automaton state
	return self.automaton_state
	
	
	# now override the step function
	LDBA.step = step.__get__(goal1_or_goal2, LDBA)
\end{lstlisting}
\begin{figure}[!h]
	\centering
	\resizebox{0.45\columnwidth}{!}{
	\begin{tikzpicture}[shorten >=1pt,node distance=3cm,on grid,auto]  
		\node[state,initial] (q_1) {$q_0$}; 
		\node[state,accepting] (q_2) [above right=of q_1] {$q_1$}; 
		\node[state,accepting] (q_3) [below right=of q_1] {$q_2$};
		\node[state] (q_4) [below right=of q_2]{$q_{-1}$};
		\path[->] 
		(q_1) edge [below] node {$~~~\varepsilon_1$} (q_2)
		(q_1) edge [loop above] node {$\neg\mathit{unsafe}$} (q_1)
		(q_1) edge [below] node {$\varepsilon_2~~~$} (q_3)
		(q_2) edge  [loop above] node {$\mathit{goal1}\wedge\neg\mathit{unsafe}$} (q_2)
		(q_2) edge node {$\neg\mathit{goal1}\vee\mathit{unsafe}$} (q_4)
		(q_3) edge [below] node {$\hspace{25mm}\neg\mathit{goal2}\vee\mathit{unsafe}$} (q_4)
		(q_1) edge node {$\mathit{unsafe}$} (q_4)
		(q_4) edge  [loop right] node {$\mathit{true}$} (q_4)
		(q_3) edge  [loop below] node {$\mathit{goal2}\wedge\neg\mathit{unsafe}$} (q_3);
	\end{tikzpicture}
	}
\caption{LDBA for the LTL specification $(\lozenge \square \mathit{goal1} \vee \lozenge \square \mathit{goal2}) \wedge \square \neg \mathit{unsafe}$.}
\label{fig:sample_ldba}
\end{figure}
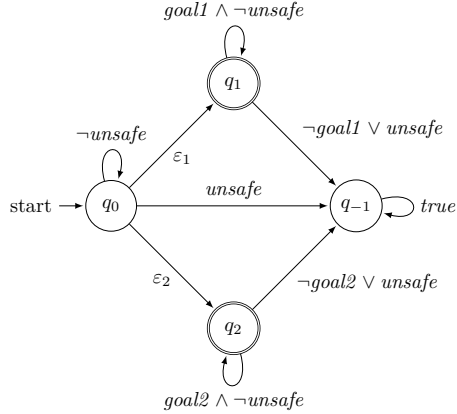

\end{itemize}

If the automaton happens to have $\varepsilon$-transitions, e.g. Fig.~\ref{fig:sample_ldba}, they have to
distinguishable, e.g.  numbered.  For instance, there exist two
$\varepsilon$-transitions in the LDBA in Fig.~\ref{fig:sample_ldba} and each is marked by an integer. 
Furthermore, the LDBA class has an attribute called
\texttt{epsilon\_transitions}, which is a dictionary to specify which states
in the automaton contain $\varepsilon$-transitions. In
Fig.~\ref{fig:sample_ldba}, only \texttt{state 0} has outgoing
$\varepsilon$-transitions and thus, the attribute
\texttt{epsilon\_transitions} in the LDBA object \texttt{goal1\_or\_goal2}
has to be set to
\begin{center}
	\texttt{goal1\_or\_goal2.epsilon\_transitions=\{0:[`epsilon\_0', `epsilon\_1']\}} 
\end{center} 

\subsection{Output Interface}

\lcrl provides the results of learning and testing as \texttt{.pkl} files. Tests are closed-loop simulations where we apply the learned policy over the MDP and observe the results.
For any selected learning algorithm, the learned model is saved as
\texttt{learned\_model.pkl} and test results as \texttt{test\_results.pkl}. 
The instruction on how to load these files is also displayed at the end of
training for ease of re-loading data and for post-processing.  Depending on
the chosen learning algorithm, \lcrl generates a number of plots to
visualise the learning progress and the testing results.  These plots are saved
in the \texttt{save\_dir} directory.  The user has the additional option to
export a generated animation of the testing progress, \lcrl prompts 
this option to the user following the completion of the test.  During the learning
phase, \lcrl displays the progress in real-time and allows the user to stop
the learning task (in an any-time fashion) and save the generated outcomes. 

\section{Experimental Evaluation}\label{sec:experiments}

\begin{figure}[!t]
	\centering
	\subfloat[]{{\hspace{-4mm}\includegraphics[width=0.39\columnwidth]{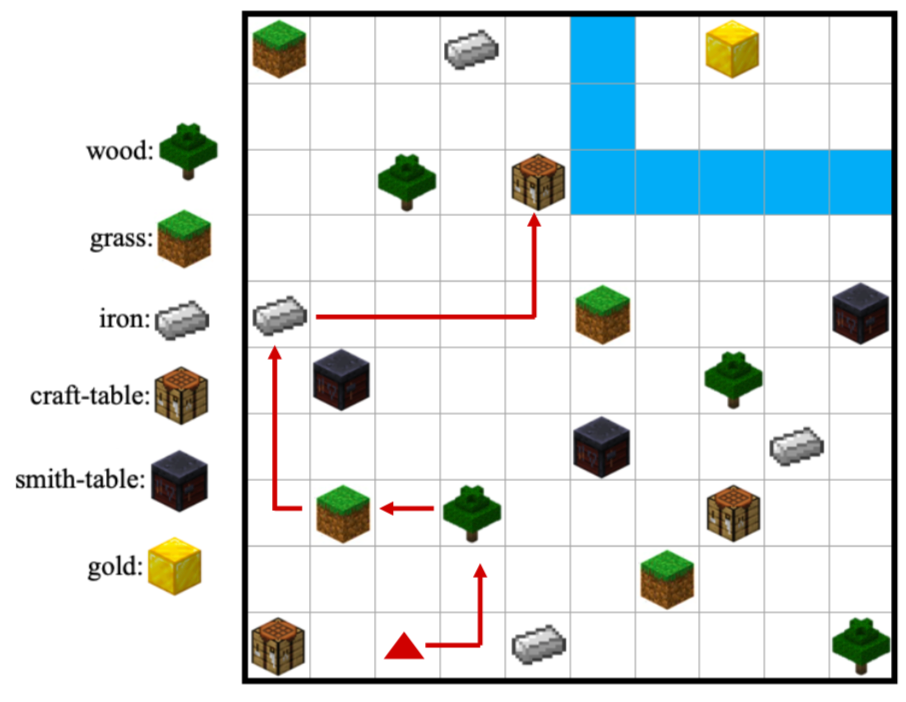} }}%
	\subfloat[]{{\includegraphics[width=0.3\columnwidth]{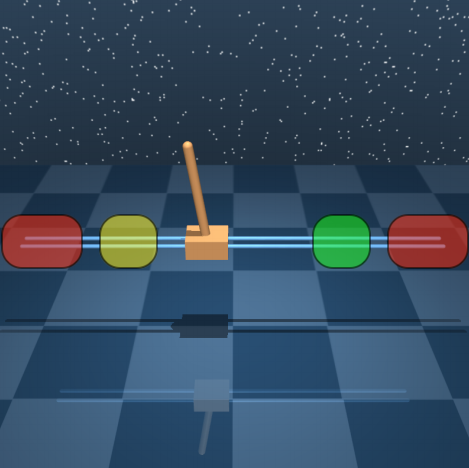} }}%
	\subfloat[]{{\includegraphics[width=0.32\columnwidth]{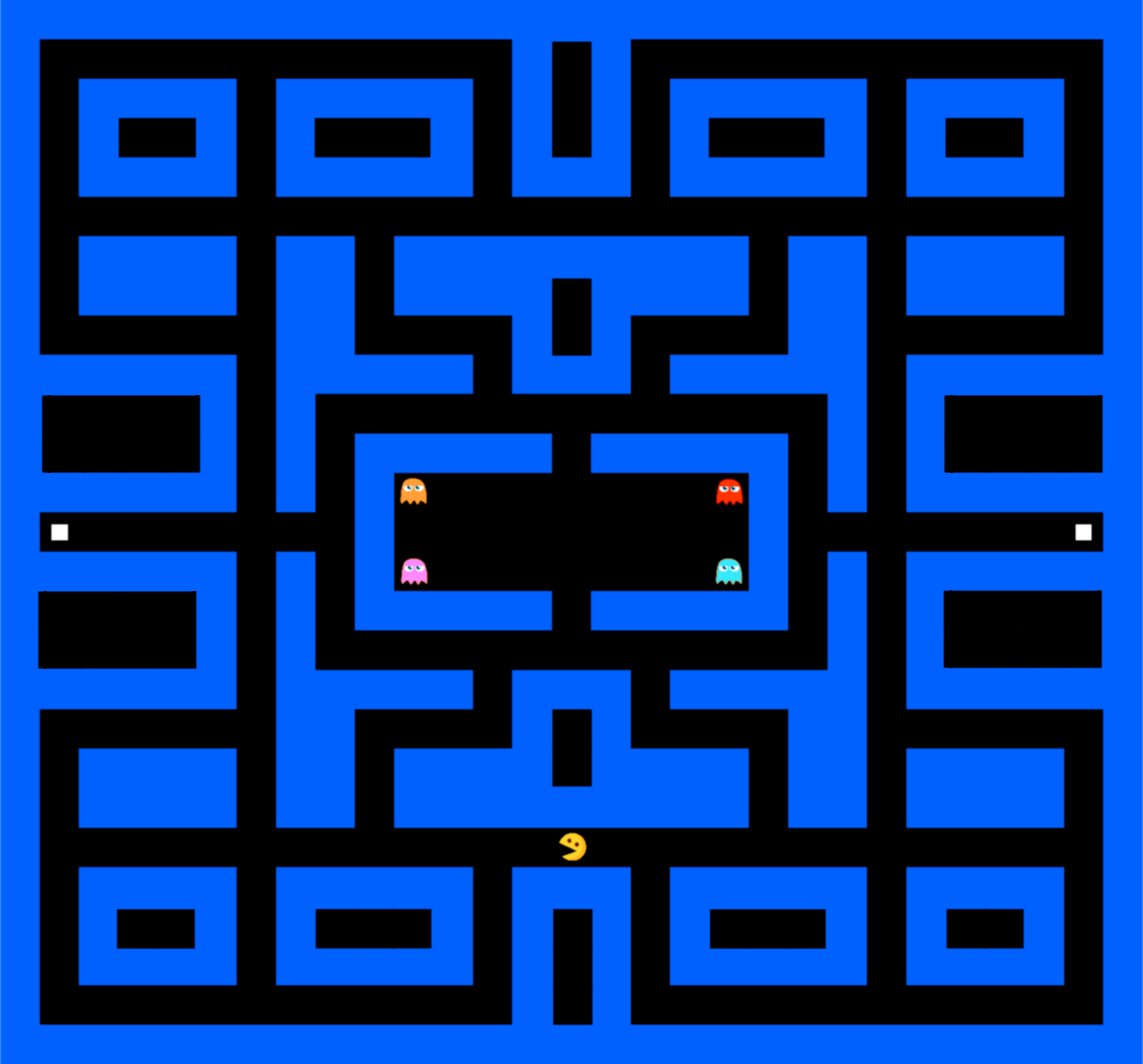} }}%
	\caption{(a) Synthesised policy by \lcrl in $\mathtt{minecraft\text{-}t3}$; (b) $\mathtt{cart\text{-}pole}$ experiment~\cite{tassa2018deepmind}; (c) $\mathtt{pacman\text{-}lrg}$ - the white square on the left is labelled as food 1 ($ f_1 $) and the one on the right as food 2 ($ f_2 $), the state of being caught by a ghost is labelled as ($ g $) and the rest of the state space is labelled as neutral ($ n $).}%
	\label{MDPcart}%
\end{figure}

\bgroup
\def\arraystretch{1}
\begin{table}[!t]
	\centering
	\caption{Learning results with \lcrl. MDP state and action space cardinalities are $|\mathcal{S}|$ and $|\mathcal{A}|$, the number of automaton states in LDBA is denoted by $|\mathcal{Q}|$, 
	the optimal action value function in the initial state is denoted by ``\lcrl $\max_a Q(s_0,a)$'', which represents the \lcrl estimation of the maximum satisfaction probability. For each experiment, the reported result includes the mean and the standard error of ten learning trials with \lcrl. 
	This probability is also calculated by the PRISM model checker \cite{prism} and, whenever the MDP model can be processed by PRISM, it is reported in column ``\texttt{max sat. prob. at $s_0$}''. 
	The closer ``\lcrl $\max_a Q(s_0,a)$'' and ``\texttt{max sat. prob. at $s_0$}'' the better. Note that for continuous-state-action MDPs the maximum satisfaction probability cannot be precisely computed by model checking tools, unless abstraction approximation techniques are applied, hence ``n/a''. Furthermore, if the MDP state (or action) space is large enough, e.g. \texttt{pacman}, the model checkers tools cannot parse the model and the model checking process times out, i.e. ``t/o''. The column ``\lcrl \texttt{conv. ep.}'' presents the episode number in which \lcrl converged. Finally, ``\crv{wall\_clocktime}'' presents the average elapsed real time needed for \lcrl to converge on a test machine. The rest of the columns provide the values
	of the hyper-parameters, as described in Table \ref{tab:hyper_parameters}.}
	\label{tab:results}
	\vspace{2mm}
	{
	\resizebox{1\textwidth}{!}{%
		%\hspace{-17mm}
		\begin{tabular}{|l|c|c|c|c|c|c|c|c|c|c|c|c|}
			\hline
			\multirow{ 2}{*}{~\texttt{experiment}} \multirow{ 2}{*}&{\texttt{MDP}} & \texttt{LDBA} & \lcrl $\max_a$& \texttt{max sat.} & \multirow{ 2}{*}{\texttt{alg.}} &{\texttt{episode\_}} & \texttt{iteration\_}& \texttt{discount\_}& \texttt{learning\_}&\texttt{\crv{wall\_clock}}\\
			& ~{$|\mathcal{S}|, |\mathcal{A}|$}~ &~{$|\mathcal{Q}|$}~ & $Q(s_0,a)$ & \texttt{prob. at} $s_0$ &&\texttt{num}& {\texttt{num\_max}} & \texttt{factor}{\Large ${}^*$}& \texttt{rate{\large ${}^\dag$}}&\crv{\texttt{time{${}^\bigstar$(min)}}}\\ \hline
			\texttt{minecraft-t1}  &  100, 5 & 3 &   0.991 $\pm$ 0.009& 1 & \texttt{`ql'} & 500 & 4000 & 0.95 & 0.9 & 0.1 \\
			\texttt{minecraft-t2}  &  100, 5 & 3 &   0.991 $\pm$ 0.009 & 1 & \texttt{`ql'} & 500 & 4000 & 0.95 & 0.9 & 0.1\\
			\texttt{minecraft-t3}  &  100, 5 & 5 &   0.993 $\pm$ 0.007 & 1 & \texttt{`ql'} & 1500 & 4000 & 0.95 & 0.9 & 0.25\\
			\texttt{minecraft-t4}  &  100, 5 & 3 &   0.991 $\pm$ 0.009 & 1 & \texttt{`ql'} & 500 & 4000 & 0.95 & 0.9 & 0.1\\
			\texttt{minecraft-t5}  &  100, 5 & 3 &   0.995 $\pm$ 0.005 & 1 & \texttt{`ql'} & 500 & 4000 & 0.95 & 0.9 & 0.1\\
			\texttt{minecraft-t6}  &  100, 5 & 4 &   0.995 $\pm$ 0.005 & 1 & \texttt{`ql'} & 1500 & 4000 & 0.95 & 0.9 & 0.25\\
			\texttt{minecraft-t7}  &  100, 5 & 5 &   0.993 $\pm$ 0.007 & 1 & \texttt{`ql'} & 1500 & 4000 & 0.95 & 0.9 & 0.5\\
			\texttt{mars-rover-1}  &$\infty$, 5& 3 & 0.991 $\pm$ 0.002& n/a & \texttt{`nfq'} & 50 & 3000 & 0.9 & 0.01 & 550\\
			\texttt{mars-rover-2}  &$\infty$, 5& 3 & 0.992 $\pm$ 0.006 & n/a & \texttt{`nfq'} & 50 & 3000 & 0.9 & 0.01 & 540\\
			\texttt{mars-rover-3}  &$\infty$, $\infty$& 3 & n/a & n/a & \texttt{`ddpg'} & 1000 & 18000 & 0.99 & 0.05 & 14\\
			\texttt{mars-rover-4}  &$\infty$, $\infty$& 3 & n/a & n/a & \texttt{`ddpg'} & 1000 & 18000 & 0.99 & 0.05 & 12\\
			\texttt{cart-pole}  &$\infty$, $\infty$& 4 & n/a & n/a & \texttt{`ddpg'} & 100 & 10000 & 0.99 & 0.02 & 1\\
			\texttt{robot-surve}   &   25, 4 & 3 &  0.994 $\pm$ 0.006 & 1 & \texttt{`ql'} & 500 & 1000 & 0.95 & 0.9 & 0.1\\
			\texttt{slp-easy-sml}  &  120, 4 & 2 &  0.974 $\pm$ 0.026 & 1 & \texttt{`ql'} & 300 & 1000 & 0.99 & 0.9 & 0.1\\
			\texttt{slp-easy-med}  &  400, 4 & 2 & 0.990 $\pm$ 0.010 & 1 & \texttt{`ql'} & 1500 & 1000 & 0.99 & 0.9 & 0.25\\
			\texttt{slp-easy-lrg}  & 1600, 4 & 2 & 0.970 $\pm$ 0.030 & 1 & \texttt{`ql'} & 2000 & 1000 & 0.99 & 0.9 & 2\\
			\texttt{slp-hard-sml}  &  120, 4 & 5 &  0.947 $\pm$ 0.039 & 1 & \texttt{`ql'} & 500 & 1000 & 0.99 & 0.9 & 1\\
			\texttt{slp-hard-med}  &  400, 4 & 5 & 0.989 $\pm$ 0.010 & 1 & \texttt{`ql'} & 4000 & 2100 & 0.99 & 0.9 & 5\\
			\texttt{slp-hard-lrg}  & 1600, 4 & 5 & 0.980 $\pm$ 0.016 & 1 & \texttt{`ql'} & 6000 & 3500 & 0.99 & 0.9 & 9\\
			\texttt{frozen-lake-1} &  120, 4 & 3 &  0.949 $\pm$ 0.050 & 0.9983 & \texttt{`ql'} & 400 & 2000 & 0.99 & 0.9 & 0.1\\
			\texttt{frozen-lake-2} &  400, 4 & 3 & 0.971 $\pm$ 0.024 & 0.9982 & \texttt{`ql'} & 2000 & 2000 & 0.99 & 0.9 & 0.5\\
			\texttt{frozen-lake-3} & 1600, 4 & 3 &  0.969 $\pm$ 0.019 & 0.9720 & \texttt{`ql'} & 5000 & 4000 & 0.99 & 0.9 & 1\\
			\texttt{frozen-lake-4} &  120, 4 & 6 &  0.846 $\pm$ 0.135 & 0.9728 & \texttt{`ql'} & 2000 & 2000 & 0.99 & 0.9 & 1\\
			\texttt{frozen-lake-5} &  400, 4 & 6 & 0.735 $\pm$ 0.235 & 0.9722 & \texttt{`ql'} & 7000 & 4000 & 0.99 & 0.9 & 2.5\\
			\texttt{frozen-lake-6} & 1600, 4 & 6 & 0.947 $\pm$ 0.011 & 0.9467 & \texttt{`ql'} & 5000 & 5000 & 0.99 & 0.9 & 9\\
			\texttt{pacman-sml} & 729,000, 5 & 6 & 0.290 $\pm$ 0.035 & t/o${}^\ddag$ & \texttt{`ql'} & 80e3 & 4000 & 0.95 & 0.9 & 1600\\
			\texttt{pacman-lrg} & 4,251,000, 5 & 6 & 0.282 $\pm$ 0.049 & t/o${}^\ddag$ & \texttt{`ql'} & 180e3 & 4000 & 0.95 & 0.9 & 3700\\
			\hline              
	\end{tabular}}
	}
	{\\\vspace{1mm} * coefficient $\eta$ in \eqref{eq:gamma_in} ~ $\dag$ learning rate $\mu$  ~ $\ddag$ timed out: too large for model-checking tools ~ ${}^\bigstar$ \crv{on a machine running macOS~11.6.5 with Intel Core i5 CPU at 2.5~GHz and with 20~GB of RAM}}
	\vspace{-4mm}
\end{table}

\begin{table}[]
	\centering
	\caption{Robustness of \lcrl performance against hyper-parameter tuning, for the \texttt{frozen-lake-1} experiment. Maximum probability of satisfaction is $99.83\%$ as calculated by PRISM (cf. Table \ref{tab:results}). The reported values are the percentages of times that execution of \lcrl final policy produced traces that satisfied the LTL property. Statistics are taken over $10$ trainings and $100$ testing for each training, namely $1000$ trials for each hyper-parameter configuration.}
	\label{tab:robustness}
	\vspace{2mm}
	\resizebox{0.75\textwidth}{!}{%
		\begin{tabular}{|c|c|c|c|c|c|}
			\hline
			\diagbox{~~$\eta$~~}{~~$\mu$~~}& 0.2 & 0.4 & 0.6 & 0.8 & 0.99 %& 0.2 & 0.4 & 0.6 & 0.8 & 0.99 
			\\ \hline
			0.2 & 92.5 $\pm$ 7.5\% & 96.7 $\pm$ 3.2\% & 91.3 $\pm$ 8.7\% & 98.8 $\pm$ 1.1\% & 94.7 $\pm$ 5.29\% %& 0 $\pm$ 0\% & 0 $\pm$ 0\% & 0 $\pm$ 0\% & 0 $\pm$ 0\% & 0 $\pm$ 0\% 
			\\ \hline
			0.4 & 98.6 $\pm$ 1.4\% & 89.5 $\pm$ 10.5\% & 94.5 $\pm$ 5.5\% & 94.5 $\pm$ 5.5\% & 99.2 $\pm$ 0.74\% %& 0 $\pm$ 0\% & 0 $\pm$ 0\% & 0 $\pm$ 0\% & 0 $\pm$ 0\% & 0 $\pm$ 0\% 
			\\ \hline
			0.6 & 99.0 $\pm$ 0.83\% & 94.5 $\pm$ 5.5\% & 93.3 $\pm$ 6.7\% & 96.4 $\pm$ 3.59\% & 93.3 $\pm$ 6.7\% %& 0 $\pm$ 0\% & 0 $\pm$ 0\% & 0 $\pm$ 0\% & 0 $\pm$ 00\% & 0 $\pm$ 0\% 
			\\ \hline
			0.8 & 95.8 $\pm$ 4.2\% & 99.5 $\pm$ 0.49\% & 99.5 $\pm$ 0.49\% & 96.9 $\pm$ 3.09\% & 97.7 $\pm$ 2.2\% %& 0$\pm$ 0\% & 0 $\pm$ 0\% & 0 $\pm$ 0\% & 0 $\pm$ 0\% & 0 $\pm$ 0\% 
			\\ \hline
			0.99 & 88.9 $\pm$ 11.09\% & 98.4 $\pm$ 1.55\% & 97.1 $\pm$ 2.31\% & 96.1 $\pm$ 3.73\% & 95.2 $\pm$ 4.79\% %& 0 $\pm$ 0\% & 0 $\pm$ 0\% & 0 $\pm$ 0\% & 0 $\pm$ 0\% & 0 $\pm$ 0\% 
			\\ \hline
			overall avg. & \multicolumn{5}{c|}{\textbf{95.676} $\pm$ \textbf{4.268}\%} %& \multicolumn{5}{c|}{\textbf{0}\%} 
			\\ \hline
		\end{tabular}%
	}
\end{table}
\egroup

We apply \lcrl on a number of case studies highlighting its features, 
performance and robustness across various environment domains and tasks.  All the
experiments are run on a standard machine, with an Intel Core i5 CPU at 2.5~GHz and with 20~GB of RAM. The experiments are listed in Table \ref{tab:results} and discussed next.   

The \texttt{minecraft} environment~\cite{pol-sketch} requires
solving challenging low-level control tasks
($\mathtt{minecraft\text{-}tX}$), and features many sequential 
%high-level
goals.  For instance, in $\mathtt{minecraft\text{-}t3}$ 
(Fig.~\ref{MDPcart}.a) the agent is tasked with collecting three items sequentially and to reach a final checkpoint, 
which is encoded as the following LTL specification: 
%
%\begin{equation*}
%\begin{aligned}
$\Diamond(\mathit{wood}\wedge\Diamond (\mathit{grass} \wedge\Diamond(\mathit{iron} \wedge\Diamond(\mathit{craft\_table}))))$,   
%\end{aligned}
%\label{minecraft_task_3_spec}
%\end{equation*}
where $\Diamond$ is the known \emph{eventually} temporal operator.

The $\mathtt{mars\text{-}rover}$ problems are realistic robotic benchmarks taken
from~\cite{lcnfq}, where the environment features continuous state and action spaces. 

The known $\mathtt{cart\text{-}pole}$ experiment (Fig.~\ref{MDPcart}.b)~\cite{tassa2018deepmind,hasanbeig2020deep,cai2021} has a task that is expressed by the LTL specification 
%
%\begin{equation*}\label{c_ltl}
$\square\lozenge y \wedge \square\lozenge g \wedge \square \neg u$, 
%\end{equation*} 
%
namely, starting the pole in upright position, 
the goal is to prevent it from falling over ($\square \neg u$, namely \emph{always not $u$}) by moving the cart, 
whilst in particular alternating between the yellow ($y$) and green ($g$) regions ($\square\lozenge y \wedge \square\lozenge g$), while avoiding the red (unsafe) parts of the track ($\square \neg u$).  

The $\mathtt{robot\text{-}surve}$ example~\cite{sadigh2014learning} has the task to repeatedly visit two regions ($A$ and $B$) in
sequence, while avoiding multiple obstacles ($C$) on the way:  
$\square\lozenge A \wedge \square\lozenge B \wedge \square \neg C$.

Environments $\mathtt{slp\text{-}easy}$ and $\mathtt{slp\text{-}hard}$ are
inspired by the widely used stochastic MDPs in~\cite[Chapter~6]{sutton}: the goal
in $\mathtt{slp\text{-}easy}$ is to reach a particular region of the state space, 
whereas the goal in $\mathtt{slp\text{-}hard}$ is to visit four distinct
regions sequentially in a given order.  

The $\mathtt{frozen\text{-}lake}$ benchmarks are adopted from the OpenAI Gym~\cite{gym}: the first three are reachability problems, whereas
the last three require sequential visits of four regions, in the presence of unsafe regions to be always avoided.    

Finally, $\mathtt{pacman\text{-}sml}$ and 
$\mathtt{pacman\text{-}lrg}$ are inspired by the well-known Atari game
Pacman, and are initialised in a tricky configuration
($\mathtt{pacman\text{-}lrg}$ as in Fig.~\ref{MDPcart}.c), 
which is likely for the agent to be caught: in order to win the game, the agent has to collect
the available tokens (food sources) without being caught by moving ghosts.  Formally, the
agent is required to choose between one of the two available foods and then
find the other one ($ \lozenge [ (f_1 \wedge \lozenge f_2) \vee (f_2 \wedge
\lozenge f_1)] $), while avoiding the ghosts ($ \square \neg g$).  
We thus feed to the agent a conjunction of these associations, as the following LTL specification: 
$
%\label{pacman_p}
\lozenge [ (f_1 \wedge \lozenge f_2) \vee (f_2 \wedge \lozenge f_1)]  \wedge \square \neg g$. 
Standard QL fails to find a policy generating satisfying traces for this experiment. 
We emphasise that the two tasks in \texttt{cart-pole} and \texttt{robot-surve} are not co-safe, namely require possibly infinite traces as witnesses.  

%The \texttt{minecraft} environment that is adopted from~\cite{pol-sketch}
%requires solving challenging low-level control tasks
%($\mathtt{minecraft\text{-}tX}$), and features highly sequential high-level
%goals.  The $\mathtt{mars\text{-}rover}$ benchmarks are taken
%from~\cite{lcnfq}, and the models feature uncountably infinite (continuous)
%state and action spaces.  The example $\mathtt{robot\text{-}surve}$ is
%adopted from~\cite{sadigh2014learning}, and the task is to visit two regions in sequence
%where there exist multiple obstacles on the way.  Models
%$\mathtt{slp\text{-}easy}$ and $\mathtt{slp\text{-}hard}$ are inspired by
%the noisy MDPs of Chapter~6 in~\cite{sutton}.  The goal in
%$\mathtt{slp\text{-}easy}$ is to reach a particular region of the MDP and
%the goal in $\mathtt{slp\text{-}hard}$ is to visit four distinct regions
%sequentially in proper order.  The $\mathtt{frozen\text{-}lake}$ benchmarks
%are similar: the first three are simple reachability problems and the last
%three require sequential visits of four regions, except that now there exist
%unsafe regions as well.  The $\mathtt{frozen\text{-}lake}$ MDPs are
%stochastic and are adopted from the OpenAI Gym~\cite{gym}.

Additionally, we have evaluated the \lcrl robustness to RL key hyper-parameter tuning, i.e. discount factor $\eta$ and learning rate $\mu$, by
training the \lcrl agent for $10$ times and testing its final policy for $100$
times.  The evaluation results and an overall rate of satisfying the given
LTL specifications are reported for the \texttt{frozen-lake-1} experiments in
Table \ref{tab:robustness}.  The statistics are taken across $10\times100$
tests, which results in $1000$ trials for each hyper-parameter
configuration.

\section{Conclusions and Extensions}

This paper presented \lcrl, a new software tool for policy synthesis with RL
under LTL and omega-regular specifications.  There is a plethora of
extensions that we are planning to explore.  In the short term, we intend
to: (1) directly interface \lcrl with automata synthesis tools such as
\texttt{OWL} \cite{owl}; (2) link \lcrl with other model checking tools such
as PRISM \cite{prism} and Storm \cite{storm}; and (3) embed more RL algorithms for policy
synthesis, so that we can tackle policy synthesis problems for more challenging environments. 
In the longer term, we plan to extend \lcrl such that (1)~it will be able to
handle other forms of temporal logic, e.g., signal temporal logic; and (2)~it will have a graphical
user-interface for the ease of interaction.

\clearpage
\bibliographystyle{plain}
\bibliography{bibliography}

\end{document}